\documentclass[letterpaper, 10 pt, conference]{ieeeconf}  
\usepackage{color}
\usepackage{graphicx}
\usepackage{booktabs}
\usepackage{multirow}
\usepackage{siunitx}
\usepackage{multicol}
\usepackage{leftidx}
\usepackage{amsmath}
\usepackage{graphicx}
\usepackage{graphics}
\usepackage{xspace}

\IEEEoverridecommandlockouts    

\makeatletter
\DeclareRobustCommand\onedot{\futurelet\@let@token\@onedot}
\def\@onedot{\ifx\@let@token.\else.\null\fi\xspace}

\def\etal{\emph{et al}\onedot}
\makeatother

\title{\LARGE \bf
Sharing Cognition: Human Gesture and Natural Language Grounding Based Planning and Navigation for Indoor Robots
}
\author{Gourav Kumar$^{*}$, Soumyadip Maity$^{*}$, Ruddra dev Roychoudhury$^{*}$, Brojeshwar Bhowmick $^{*}$ 
\thanks{$^{*}$ TCS Research and Innovation Labs, Kolkata, India}
}

\begin{document}

\maketitle

\begin{abstract}
Cooperation among humans makes it easy to execute tasks and navigate seamlessly even in unknown scenarios. With our individual knowledge and collective cognition skills, we can reason about and perform well in unforeseen situations and environments. To achieve a similar potential for a robot navigating among humans and interacting with them, it is crucial for it to acquire the ability for easy, efficient and natural ways of communication and cognition sharing with humans. In this work, we aim to exploit human gestures which is known to be the most prominent modality of communication after the speech. We demonstrate how the incorporation of gestures for communicating spatial understanding can be achieved in a very simple yet effective way using a robot having the vision and listening capability. This shows a big advantage over using only Vision and Language-based Navigation, Language Grounding or Human-Robot Interaction in a task requiring the development of cognition and indoor navigation. We adapt the state-of-the-art modules of Language Grounding and Human-Robot Interaction to demonstrate a novel system pipeline in real-world environments on a Telepresence robot for performing a set of challenging tasks. To the best of our knowledge, this is the first pipeline to couple the fields of HRI and language grounding in an indoor environment to demonstrate autonomous navigation.

\end{abstract}
\section{Introduction}

Since the inception of robotics, navigation in an environment built as per human convenience has been both challenging as well as of high significance. Leveraging the collective cognition of the social environment around us for conducting any task is a  capability that makes it possible to reliably perform tasks to which the group has not been exposed apriori. To inculcate this capability in robots for navigation in an unknown environment, we need to have a way to communicate with humans naturally.\\

People of all ages, culture and background gestures when they speak for navigational instructions. The use of pointing gesture is one of the first ways in which people communicate with the world \cite{pointing_and_language1}. It is a foundational building block of human communication because it is used during the early phases of language development in combination with speech, in order to name objects, indicating a developmentally early correspondence between word and gesture \cite{gesture_and_teaching}. These hand movements are so natural and pervasive that researchers across many fields, from linguistics to psychology to neuroscience, have claimed that the two modalities form an integrated system of meaning during language production and comprehension \cite{gesture_in_cognition}. It is well known that pointing at a reference object is a much faster and more convenient method than describing it verbally. As the need of gesture for directional and contextual sense while speaking about the spatial environment around us, it becomes very crucial to incorporate both the modalities of communication for this task. The need to understand and exploit this type of communication leads to the combination of the two major fields - Human-Robot Interaction and Visual Language Grounding.\\
\begin{figure}[!ht]
\includegraphics[width=8.9cm, height=5.5cm]{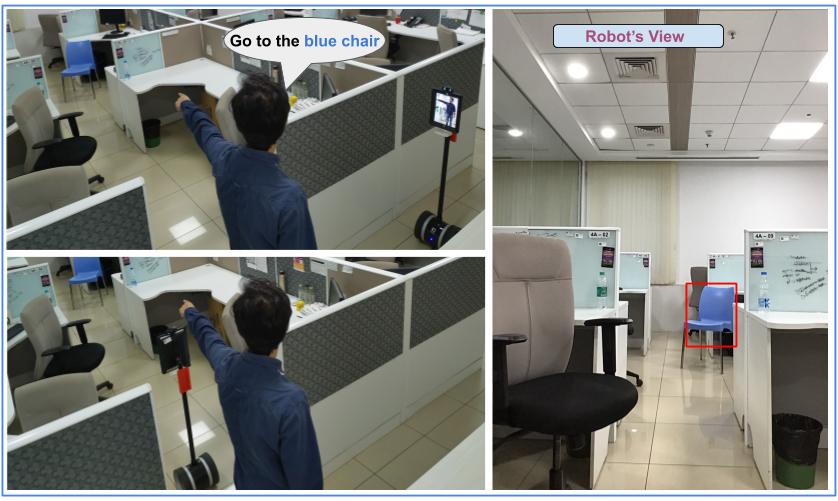}
\label{fig:teaser} 
\caption{\textbf{Cognition Sharing:} Replying to the query of robot about a location environment, a human guide communicates the information using natural language and body gesture while the robot observes him/her.}
\vspace{-0.5em}
\end{figure}

    As these two senses are uncorrelated at the level of raw inputs, it becomes a matter of utmost importance to learn the correlation between these two to have a more robust way of communication and context-aware understanding of the world. Recent works in Vision and Language based Navigation \cite{talk_to_the_vehicle}, \cite{Behavioural} work on only on the language commands. Providing the natural language commands to a robot poses few challenges for a human bystander from whom the robot seeks help. Having a depth of the scene computed from RGB \cite{broj1} or using both RGB and IMU \cite{broj2}. \cite{broj3} may help but still associating the depth an d language is a challenge. \textit{Firstly}, communicating the exact turn angle towards the goal only through natural language command is very difficult for any person, especially in cases where the robot may be oriented in any direction(towards the person in our case) hence predefined discrete actions may not be of much help in many of the indoor scenarios. \textit{Secondly}, the natural language command which is supposed to be given in the robot-centric frame, adds an extra cognitive overhead on the human instructor making the communication unnatural and inconvenient for him. We in this work intend to address these issues by transferring this cognitive load from a human to the seeker robot through our novel pipeline which computes the relative frame transformation along with understanding the navigational gesture commands simplifying the directions and easing the task of the instructor.\\
Navigational gestures communicate spatial information through gestures. Gestures come in four different types- iconic, deictics, metaphoric and beat with only iconic and deictics employed in navigation cues. Deictics are pointing gestures used by speakers to orient the listener in the referential space. We use deictics as the way of communication with the robot. Using a Kinect like sensor to understand the human state is relatively easier and have been used in many applications \cite{broj4} but in this work our robot does not have such explicit sensor and hence makes the problem harder. \\

In the area of Human-Robot Interaction(HRI), people have explored various combinations of sensor modalities to have a better communication and interaction between the humans and the robot at hand, but, the most natural way of communication comes to a person when there is no prerequisite to have a dedicated sensors/hardware mounted on them or in their immediate surrounding. This condition also makes the task of induction of robots in the human working environment smooth.\\
The reasons behind the need for gesture-based feedback are:
\begin{itemize}
\item Hand and body gesture is one of the most prominent modalities of communication and comes naturally for people of all ages, cultures, and backgrounds.

\item Current state-of-the-art Vision and Language based Navigation frameworks are suitable for navigation in structured outdoor driving scenarios or in scenarios requiring only discrete set of motions in indoor environments but makes the task complex for moving around in indoor environment which requires to move to any angle in the continuous 360 degree space.

\item In tasks like navigation, localization, and mapping pertaining to robot in an indoor scenario where multiple instances of the same object are frequently present, this leads to ambiguity in decision making. In such scenarios unique identifications of objects is crucial for completing the task. Also in situations like occluded perspective for the region of interest, one-shot determination of the goal is difficult. Hence the gesture-based feedback has two-step advantage, firstly in achieving an accurate shared perspective for better language grounding and secondly, to address the cases of ambiguity occurring due to the cases of multiple instances of the same object in the scene.

\item It can act as a sanity check for the human gesture-based region of interest proposal by evaluating the relevance between the spoken sentence and the pointing region and hence can be extended to request clarification/correction from the human agent making the pipeline more robust to human errors.

\end{itemize}

 We propose to improve as well as combine these two fields of HRI and Language Grounding such that it can complement each other and gives us an ability to address a more difficult task than that can be handled by the current state-of-the-art in a more natural and human centric way. In a summary we intend the following contributions:
 {
\begin{enumerate}

\item In this work, we propose to bring forward the important concept of cognition sharing in a more realistic and complex scenario for human-robot interaction to enable the robot to exploit the opportunity of being among people who are more familiar with its surroundings.

 \item We present a novel system pipeline of gesture and language based navigation which mitigates the issue with sharing visual perspective between human and robot in indoor narrow and cluttered environments. This issue is critical as the 360 degree view between any two locations turns out to be very different in narrow indoor spaces even if the two locations are nearby. This issue impedes one step sharing of spacial understanding between a ground robot an a person. 
 
 \item We demonstrate a complete working system that can navigate autonomously in 2D occupancy mapped indoor environment based on its own cognition and human language and gesture commands using a single monocular camera on an indoor Telepresence robot.
 
 \item We make the task of gesture-based commanding by breaking it down into pointing direction estimation to attain a shared visual perspective as an intermediate goal followed by language grounding on the shared perspective to estimate the final goal. This makes the system to be robust for any pointing direction and the goal need not be in the initial view of the robot.  
 
 To the best of our knowledge, this is the first pipeline to couple these areas of HRI and language grounding in an indoor environment to demonstrate autonomous navigation in a more human centric way.
 
\end{enumerate}
}

\section{Related Work}
 As this work combines multiple dominant fields, in this section we refer only to the most relevant in HRI and language grounding.

 \begin{figure*}[t]
\begin{center}
\includegraphics[width=18cm, height=8.8cm] {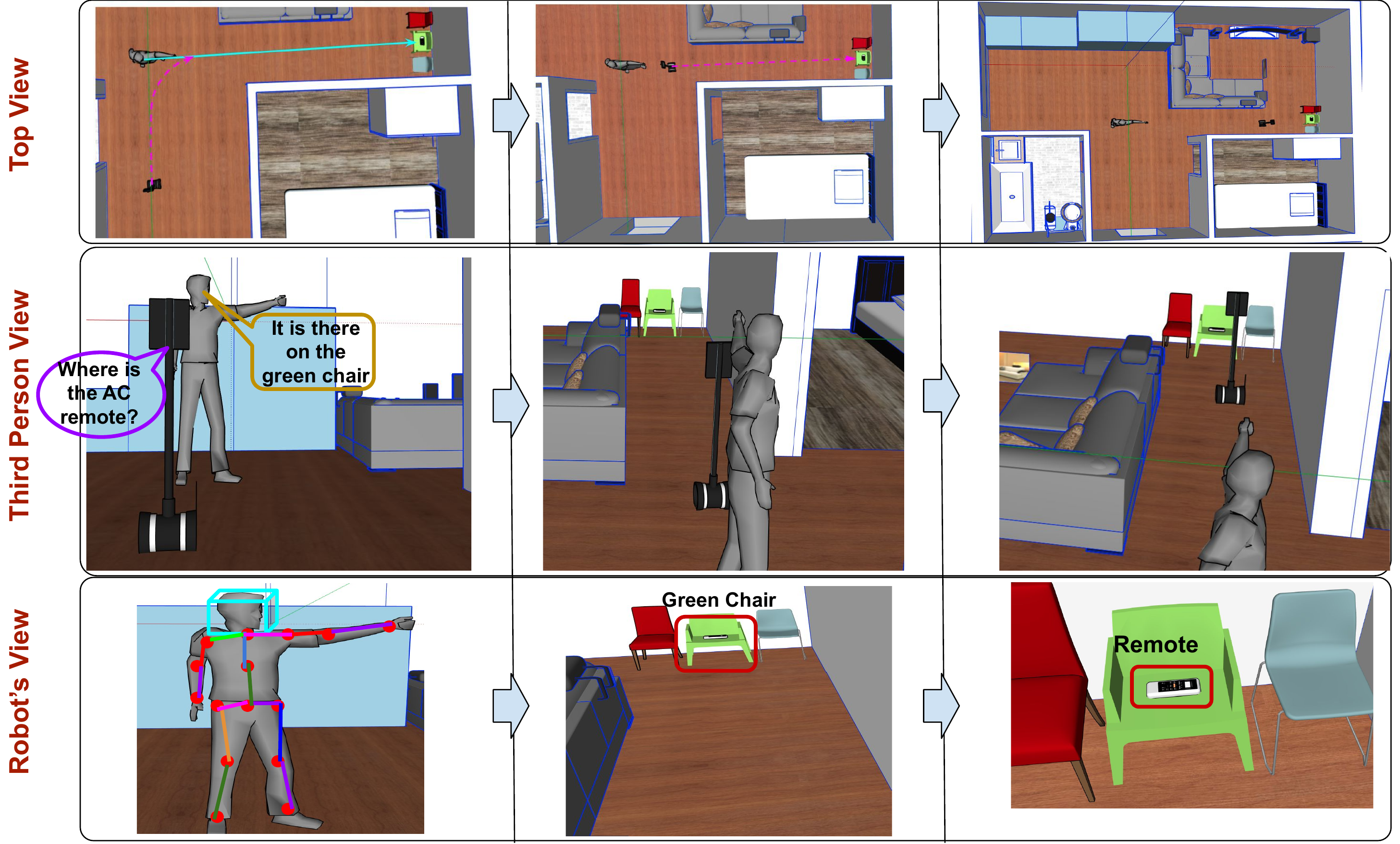}\\
\caption{ \textbf{A Scenario:} This figure illustrates a use case scenario where the robot is trying to find an AC remote. The top row shows the overall perspective of the environment in which the robot motion planning and human pointing is shown in the global perspective. The second row shows the third-person perspective of the interaction between the robot and its immediate environment. The third row shows the robot's view in which the human pose estimation and language grounding modules run and determines the robot motion plan. The figure presents the occurrence of events in order from left to right.}
\label{fig:scenario}
\end{center}
\end{figure*}

\subsubsection{Multimodal Interfaces in HRI}
 As evident from the significance of multimodal interfaces discussed in the last section, many groups around the world have made attempts to combine multiple sensor modalities for Human-Robot-Interaction. The earliest attempt along this line was presented in \cite{put_that_there} and is considered to be the forerunner of multimodal interfaces for HRI where gesture and speech are used control events on a large raster screen in a controlled environment. Most of the works attempted till now are listed in \cite{Multimodal_interaction_survey1} and \cite{Multimodal_interaction_survey2}. Despite multiple attempts with a different combinations, there still exist various challenges to be solved as can be seen in \cite{Multimodal_interaction_review}. As we focus on the two dominant modalities of speech and gesture in this work, we will confine ourselves along this line. Though the gesture and speech based interaction with different types of robot is being attempted from a long time, the capability and performance of these systems kept on improving as per evolving state-of-the-arts technology. Several attempts have been made to combine the gesture and speech in the domain of Assistive Robotics to have a more human like interaction between people and the robot. works like \cite{point_and_command}, \cite{VGPN}, \cite{SVP}, \cite{Pointing_gesture_TOF_camera}, \cite{2Stage_HMM} and \cite{Caesar} all focus on the same crucial task of human gesture based commanding a robot to perform some task but all of them use depth information using the time-of-flight sensor to estimate the human pose. This requirement of a specialized hardware makes the utility of these systems limited unlike our approach where we use only a single monocular image in our pipeline to perform the end-to-end task making it deployable on most of off the shelf robots available in the market. We use 3D human pose estimation network \cite{3D_pose} along with the face pose detection network \cite{openface} to get the accurate pose of the object or the region of the surrounding being referred to.

\subsubsection{Language Grounding}
 Object localization and modeling correspondence between language and vision are two fundamental problems of computer vision. The first problem is a major research topic from the initial years, whereas the interest on the second one has grown rapidly with the introduction of image search engines from the end of the nineteenth century. However, Barnard \etal \cite{words&pictures} were first to associate input words based on the joint distribution of image regions and words to localize objects. A significant amount of work \cite{seg&annotation, grounded_attributes, Visual_semantic_embedding, text2Image_conference} has enhanced the state-of-the-art in this following years. Recent work \cite{lg45, lg40, lg19, lg7} to co-relate a phrase with an entity of the image is termed as phrase-grounding or language-grounding, where object localization is performed based on the query phrase. While few of these works \cite{Sadhu_2019_ICCV, lg49, lg7, lg9, lg34, lg45} are supervised and use attention mechanism to associate textual and visual cues, few other approaches use supervised \cite{lg6} and unsupervised learning \cite{lg50}. Some useful datasets like Flickr30k\cite{lg8}, MSCOCO\cite{lg11}, ReferIt\cite{lg20}, Visual Genome\cite{lg44} have significant contributions to the advancement of language grounding. As an advancement, few latest works \cite{ lgh4, lgh36, lgh37} focus on modeling the association between vision and sound, but still need improvements to enhance the performance. 
   
   In our experiment, we focus on the indoor navigation of an autonomous ground vehicle. The input to the language grounding system is the natural language audio by the human instructor. The system proposed by Harwath \etal  is the state-of-the-art algorithm to fetch object regions from a given natural language audio phrase, but due to the lack of proper indoor datasets performs poorly in the indoor environment and thus measurably fails and hence cannot be used in our pipeline. We choose the method proposed by Sadhu \etal \cite{Sadhu_2019_ICCV} for their better performance in unseen scenarios. This work demonstrates a good performance with zero-shot grounding (localizing novel “unseen” objects) while maintaining the simplicity of their model. In indoor environment, we need to localize a wide variety of different indoor objects and it's not possible to add all of them in the training set. Zero-shot grounding really helps us to account for those novel objects.

\section{Problem Statement}
To demonstrate the high level cognition transfer capability, we choose an off-the-shelf \  Double2\footnote{https://www.doublerobotics.com/} teleprence robot moving in an indoor office environment of which we have a prebuilt occupancy map. We assume the environment is static with humans as the only dynamic agents apart from the robot itself. It is also assumed that humans move in such a way as to avoid colliding with the robot. Navigation between a group of people moving in the indoor environment can be an interesting extension for the future.\\
The double robot has a reliable odometry with a microphone and two monocular RGB cameras out of which we use one. The Double robot also has a display as can be seen in figure \ref{fig:scenario} and the robot as a whole communicates with an edge device with a sufficient compute power over the Wi-Fi to run the heavy compute of the pipeline. As the data needed to be transferred over the Wi-Fi is only an audio clip and few image frames, the transmission does not incur any significant delay. In the following subsections we discuss about each of the major components of the pipeline in detail.

\section{Methodology}
\begin{figure}[ht!]
\includegraphics[width=8cm, height=5cm] {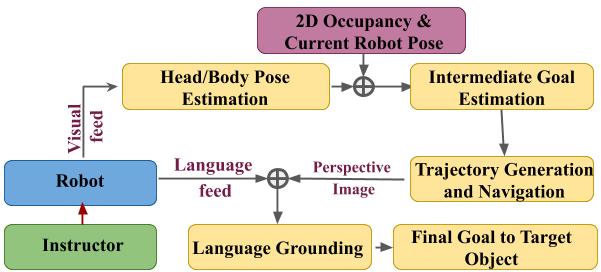}\\
\caption{\textbf{Flow diagram:} The overall behaviour flow diagram of the proposed pipeline} 
\label{fig:flow_diagram}
\end{figure}

In this section, we will go into details of each module of the pipeline. The scope of this section is also focused towards the rationale behind using the module and the adaptations made to improve the performance of the overall pipeline. The overall flow diagram of the system pipeline is illustrated in figure \ref{fig:flow_diagram}. The pipeline starts with either a robot asking a question for directional assistance or an instructor commanding the robot to perform some task. Once the user gives the speech command, the recorded voice command is passed to \textit{google cloud}'s speech-to-text\footnote{https://cloud.google.com/speech-to-text} web application to parse the voice command and generate text phrase. This processing is done while the robot navigates from starting position to the intermediate goal point and the parsed text phrase is passed on with the shared perspective image to the language grounding module.

\subsection{Pointing Direction Estimation}\label{pointing_direction_estimation}
The first crucial component of the pipeline is the estimation of pointing direction based on the gesture of a person in the field of view of the robot. The complete body gesture is obtained using these submodules: 
\subsubsection{Body Skeleton based Gesture Estimation}\label{3D_human_pose} 
\begin{figure}[ht!]
\includegraphics[width=8cm, height=5cm]{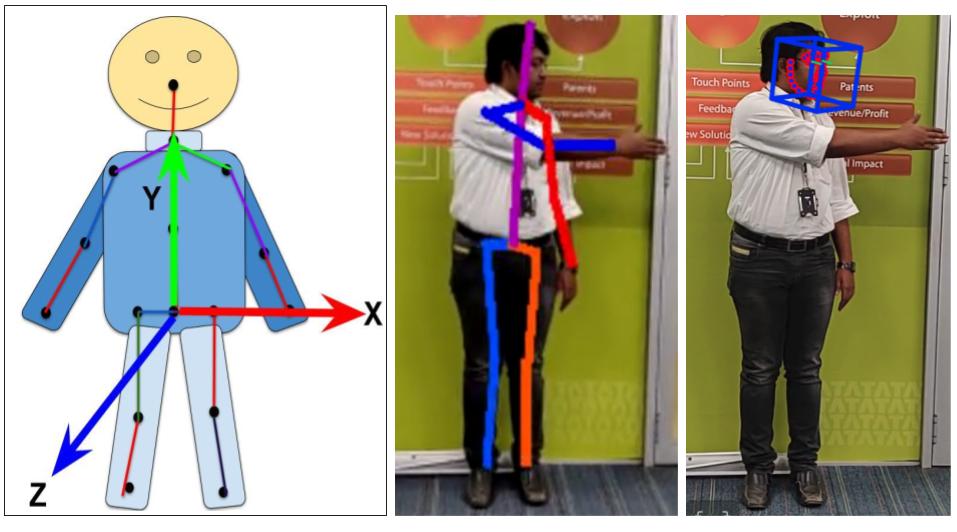}\\
\caption{\textbf{Frames} (left to right): The body frame coordinate axes convention, visualization of 3D body pose prediction and head pose prediction as described in section \ref{pointing_direction_estimation} }
\label{fig:frame_transfer}
\end{figure}

As the overall pipeline is based on perception through a monocular camera, we adapt a state-of-the-art deep neural network based 3D human pose estimation \cite{3D_pose}. The predicted 3D pose is used to initialize the human-centric coordinate frame with the origin at the pelvic joint. The X and Y axes are defined from the origin along the left hip joint and towards the spinal axis respectively as seen in \ref{fig:frame_transfer}. As the 3D pose prediction is done from a monocular image, the predicted pose is in some non-metric scale. The predicted pose is scaled by a scale factor after estimating the depth of the instructor as described in section \ref{mono_depth}.
\subsubsection{Head Pose based Gaze Direction
Estimation}\label{head_pose}
As can be seen in the state-of-the-art 3D human pose estimation works, the 3D pose prediction has an average error in the range of 35-50 mm. This error in the pose prediction makes the accurate pose estimation and hence the pointing angle uncertain. This error in pointing direction even by a small magnitude can scale up to a large error margin if the object/region of interest is at a large distance from the instructor. To make the pipeline robust and reliable in all types of scenario, it was essential to have an alternate way of estimating the pointing direction.\\
Taking the natural human behaviour into account, it does not seem wrong to assume that while speaking in the context of any spatial information in their neighbourhood, people tend to also have a quick look towards the region of interest. This may be to reconfirm the correctness of their hand gesture or to also take the attention of the listener to the region in context. This is because when two people are talking, they tend to focus mostly on each other's facial region.\\
Taking into account this aspect of human behaviour, face pose becomes the most relevant data to be fused with the human body pose. We in this approach adapt the work by Biswas \etal \cite{openface} for estimating the gaze direction while talking. Fusion of body gesture-based direction along with the human face pose provides us a very stable and robust pointing direction estimation. \\
As Pointing gesture and head pose are both estimated in the camera frame of reference. It will only require additional information of translation between the person and the robot to define the transformation between the robot's camera and the person's body frame. Also since we know the robot's body frame w.r.t. its camera frame and also w.r.t. the local map frame, the pointing direction can be directly transformed to the local map for trajectory planning. 
\subsubsection{Geometric Monocular Depth Estimation}\label{mono_depth}
As all the modules operate on a monocular image, it becomes a matter of utmost importance to estimate a scale factor which can be used to scale up all the measurements which in turn can be used to compute the transformation between the robot and the instructor for accurate transfer of the instructed goal to the robot's frame of reference. Assuming the robot and the instructor both always stand on the same leveled floor and the point where the instructor's feet touches the floor is visible in the robot's view, we can quite accurately estimate the distance between the person and the robot. Assuming a zero pitch angle of the camera w.r.t. the ground plane and given the height of the robot camera above the ground ${H}_c$, robot's camera intrinsic calibration parameters \textbf{K} = ($f_x$,$f_y$,$c_x$,$c_y$) where,$f_x$, $f_y$, $c_x$ and $c_y$ are focal lengths and camera center parameters of the camera respectively. We can compute the distance between the robot and the person \textit{d}, given the bottom of the bounding box of person detection in the image(if the contact point between the foot and the floor is visible) as \textbf{b} = ($b_x$,$b_y$).
Approximating the camera with pinhole model as in \cite{monocular_height}, we can write:

\begin{equation}\label{equation_monodepth}
\begin{split}
d = \frac{f_y{H}_c}{\Delta y_b}
    \\
where, \Delta y_b = |b_y - c_y|
\end{split}{}
\end{equation}{}

 Having the rotation from section \ref{3D_human_pose} and \ref{head_pose} and the translation information from section \ref{mono_depth}, the transformation between the robot and the instructor can be obtained. Using this the goal point generated in the instructor's reference frame can be transformed into the robot's frame.
\subsection{Shared Visual Perspective}\label{SVP}
The next step in the pipeline is to generate an intermediate goal point based on the gesture-based pointing direction estimation described in section \ref{pointing_direction_estimation}. The Intermediate goal is generated with the aim to generate the shared visual perspective of the instructor and the robot which is further used by our \textit{language grounding} \ref{language_grounding} module to generate the final goal based on the natural language instruction given by the instructor. The two sub-module required to achieve this intermediate step are:
\subsubsection{Intermediate Goal Estimation}\label{intermediate_goal}
The main objective to have an intermediate goal is to have an accurate estimate of the context in which the instructor has generated the natural language. In constrained indoor spaces, it is quite usual that the perspective of one view is occluded for the other view. This is also possible in the case of instructor's and robot's perspective as can be seen in figures \ref{fig:teaser} and \ref{fig:scenario}. To avoid any confusion for the \textit{language grounding} module, an intermediate goal is generated based on the instructor's pointing direction. The intermediate goal is chosen to be a pose on the pre-built 2D occupancy map at a distance on 1 metre from the instructor on the ray pointing towards the object/region of interest. This intermediate goal is passed on to the next sub-module \ref{navigation} for robot's navigation.
\subsubsection{Trajectory Generation and Navigation}\label{navigation}
Pre-built 2D occupancy map of the environment and the robot's odometry is used to plan and execute the control commands to reach the intermediate goal pose. As it is assumed that the environment is static, a collision-free smooth path is generated from the current robot's pose to the desired intermediate pose using the \textit{ROS movebase} local planner \cite{ros_teb}. 
Once the intermediate goal is reached, the image of the current perspective which is also the shared perspective is passed on to the language grounding module 

\subsection{Language Grounding}\label{language_grounding}
From the intermediate goal point, the language grounding method is called with the robot captured image and the natural language text from the instructor. We choose the language grounding method proposed by Sadhu \etal \cite{Sadhu_2019_ICCV} for it's suitability to our application. The authors replace the traditional two-stage object classifiers by their novel zero-shot single-stage network ZSGNet to incorporate Zero-Shot Grounding. Given dense region proposals, this network predicts the matching region and compact it by regression. The pipeline uses a language model for encoding the query text and a visual module to produce the feature-maps of the input image in different resolutions. ZSGNet takes the language features and the visual feature-maps as input and predicts the box-proposal and classification score as output. We experimentally choose a threshold 0.60 on the classification score to predict the presence of the query object in the scene.
\begin{figure}[ht!]
\includegraphics[width=8cm, height=5.5cm]{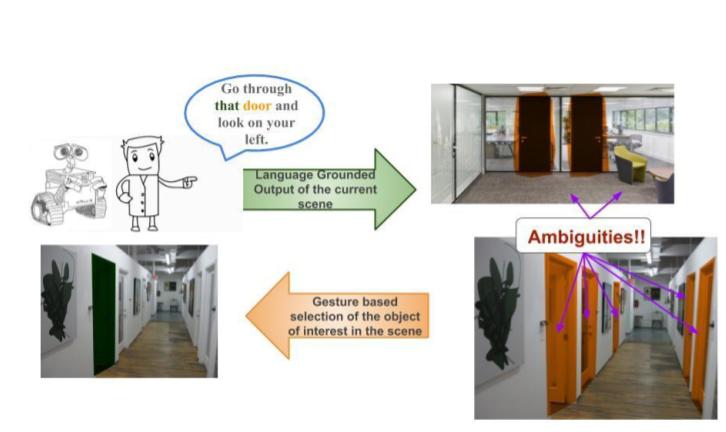}\\
\caption{\textbf{Ambiguities}: A scenario showing the ambiguity that may arise in the result of language grounding when more than one instances of object of interest are present in the scene. }
\label{fig:ambiguities}
\end{figure}
In the aspect of object localization, indoor and outdoor environments are dissimilar due to different types of objects and different nature of light sources. To make the method working in the indoor environment, we train it with a custom dataset prepared by us. We choose the MIT indoor dataset named \textit{'Indoor Scene Recognition'}  as it contains more than 15K indoor images with object annotations. Apart from this dataset, we sub-sample the indoor images from Visual Genome \cite{lg44} with region descriptions. We merge both the data and train for a better phrase to image localization. 

\subsection{Final Goal to Target Object}
Once the \textit{language grounding} module gives us the region proposals(one or more if multiple similar objects present as shown in figure \ref{fig:ambiguities}), the ambiguity in the object/region of interest is resolved by pointing direction. Once we have the detection of the object of interest in the scene, we track the object while moving in the pointing direction till either there is no further navigable space available in that direction in the map or the object gets out of the view of the robot.
\section{Experimental Setup}
To test the performance of our pipeline in the real world situations, we demonstrate the working of the system on two different types of indoor scenarios as explained in the below sections. Complete experimental video of these sections are provided in the supplementary material along with this paper. The experiment presented in the material provides an intuitive demonstration of the approach.
\begin{figure}[ht!]
\includegraphics[width=9cm, height=6cm] {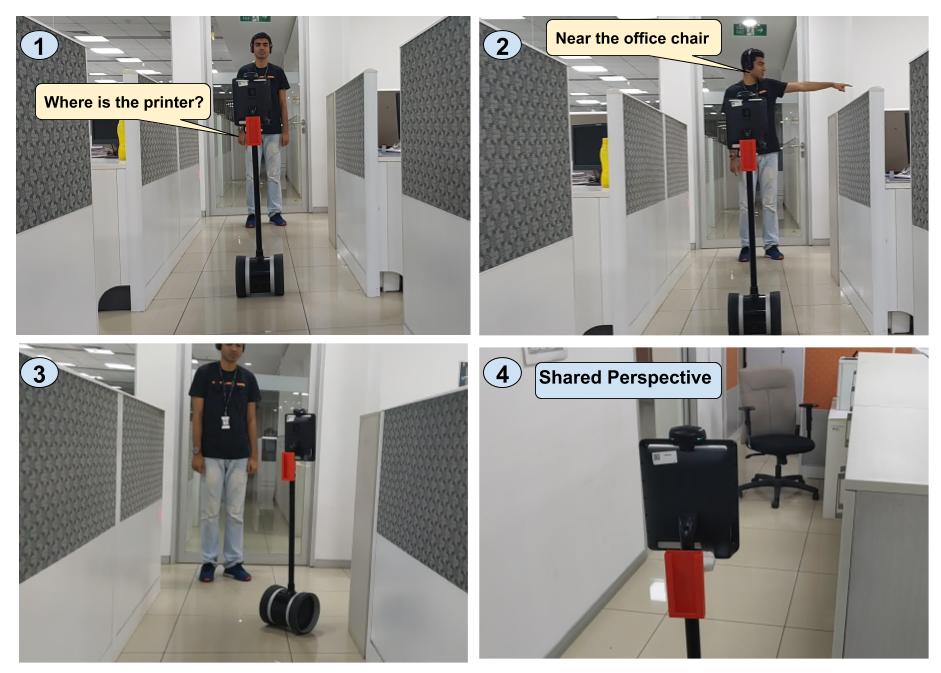}\\
\caption{\textbf{Occluded Perspective:} A typical use case for gesture based instruction to an object lying in the occluded perspective of the robot.} 
\label{fig:occluded_perspective}
\end{figure}
\subsection{Occluded Shared Perspective}
This setup is specially chosen to demonstrate the significance of intermediate goal based pipeline to exploit the concept of shared visual perspective. The setup used here is similar to what described in the figure \ref{fig:occluded_perspective} and is shown in figure \ref{fig:teaser}. In this scenario the instructor is commanding the robot to go to a chair which is not in the current visual perspective of the robot. The robot estimates the pointing direction using head/body pose estimation as described in sections \ref{head_pose}, \ref{3D_human_pose} and \ref{mono_depth} to generate the intermediate goal for the shared visual perspective. On reaching the intermediate goal, the robot sees more than one instance of object of chair class in the pointed direction. Language grounding module as described in section \ref{language_grounding} is used to resolve the ambiguity based on the current perspective image and the object attribute mentioned in the language command. The robot then moves towards the chair of interest till the complete chair is in the view and there is navigable space between the chair and the robot.
\subsection{ Referring Third Person for Guidance}
\begin{figure}[ht!]
\includegraphics[width=8cm, height=5cm]{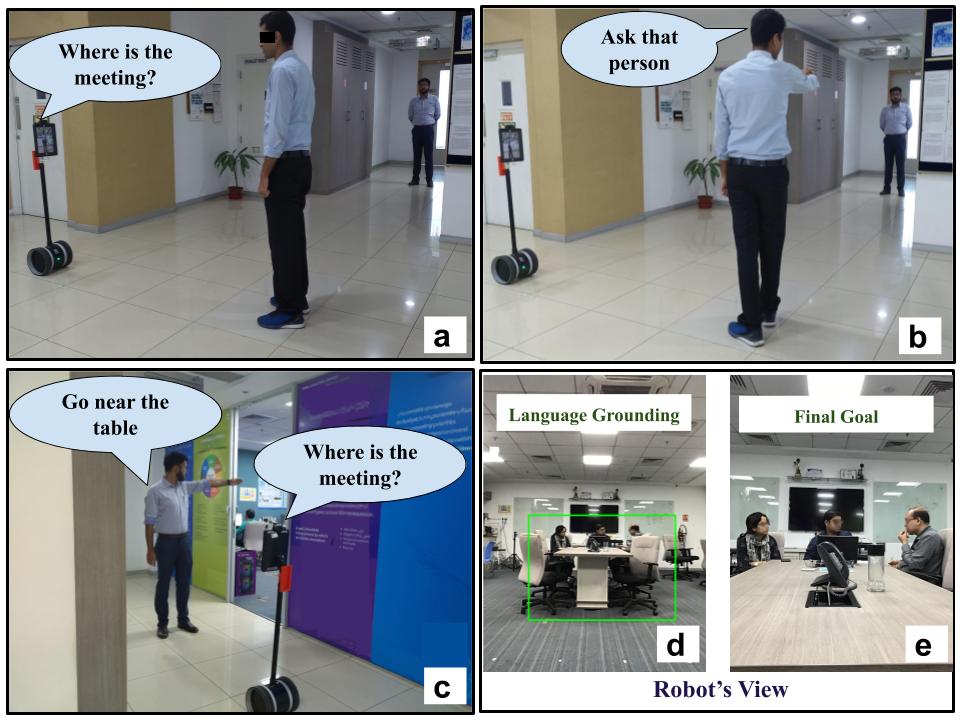}\\
\caption{\textbf{Referring Third Person for Guidance} (a) The robot approaches the first person for direction to the conference (b) The first instructor refers the robot to seek help from a third person standing nearby (c) The third person directs the person to the place where the meeting is going on (d) The shared perspective from the robot's view and the language grounding result (e) The robot's view of the final destination.}
\label{fig:third_person_reference}
\end{figure}

This experimental setup is chosen to demonstrate the scalability of the pipeline for larger task where the entire pipeline is used more than once for a single task. In this scenario person whom the robot seek help from regarding the direction refers it to a third person to seek help. The robot seeks the referred person for help regarding the directions and obtains the goal based on his/her speech and gesture command as shown in the figure \ref{fig:third_person_reference}. 

\section{Results}
This work proposes a concept of cognition sharing and a novel system pipeline to demonstrate the utility of this system in various real-world situations. As this pipeline is composed of adaptations of various open-source modules which are current state-of-the-art, it is not of much significance to evaluate them individually for their performance. Also due to the lack of any data on the performance of similar pipeline or any standard metric on which the performance can be evaluated, we report the validity of our pipeline by mostly qualitative aspects of the experiments which make this pipeline a better approach over existing approaches like \cite{SVP} and \cite{VGPN}.
\subsection{Qualitative Analysis of the Experiments}
This system pipeline takes inspiration from the fundamental behaviour of human communication and tries to replicate it for human-robot interaction. Addressing the challenges faced by each module  The biggest advantage of this approach is to improve on the ease of communication on the part of the human instructor and also to reduce the causes of failures and ambiguity on the part of robot goal finding. This is achieved by adapting the concept of language grounding on the shared visual perspective through a gesture-based intermediate goal generation. This gives the system a more robust approach and user-friendly operation, in turn, gives better performance than other state-of-the-art approaches.
\subsubsection{Limitations}
Although conceptually, this pipeline is supposed to work in most of the cases but in real-world experiments we observed that the system fails because of one of the following reasons:
\textit{Firstly}, the language grounding framework is trained on our own prepared indoor dataset, the trained network performance has a scope of improvement based on more effort on dataset preparation and fine-tuning. \textit{Secondly}, for accurate scale determination, point of contact between the person's foot and the floor needs to be in the field of view of the robot while giving instruction. \textit{Thirdly}, the pointing direction estimation module is unsuccessful where the instructor is so far away from the robot that the face of the instructor appears small which leads to failure of face pose estimation module. Although this situation appears very rarely in real-life cases as people asking help tend to come close to the bystander and so does the robot in our case.   
\section{Conclusion}
We propose a novel system framework that leverages all the recent developments in the fields of human pose prediction and language grounding to solve a more realistic and challenging problem in the domain of Human-Robot Interaction. The fundamental task of human-robot cognition sharing needs a few finer aspects of communication to mature up to reach the stage equivalent to human-human cognition sharing. Though there are few challenges left to address, this work intends to provide the foundational framework needed to address this problem as a whole. Though this work has been presented in a form to address only the indoor scenario, most of the modules used in this work are agnostic to the surrounding environment and the platform.

\bibliographystyle{ieeetr}
\bibliography{citations.bib}


\end{document}